\documentclass[a4paper]{article}

\usepackage{INTERSPEECH2020}

\title{Convolutional Neural Network-Based Age Estimation Using B-Mode Ultrasound Tongue Image}
\name{Kele Xu$^{1,2}$, Tam\'as G\'abor Csap\'o$^{3,4}$, Ming Feng $^{5}$}
\address{
	$^1$ National Key Lab of Parallel and Distributed Processing, National University of Defense Technology, Changsha, China\\
	$^2$ School of Computer, National University of Defense Technology, Changsha, China \\
  	$^3$ Department of Telecommunications and Media Informatics, \\
	Budapest University of Technology and Economics, Budapest, Hungary \\
	$^4$ MTA-ELTE Lend\"ulet Lingual Articulation Research Group, Budapest, Hungary\\
	$^5$ Tongji University, Shanghai, China
}
\email{kelele.xu@gmail.com}

\begin{document}

\maketitle
\begin{abstract}
Ultrasound tongue imaging is widely used for speech production research, and it has attracted increasing attention as its potential applications seem to be evident in many different fields, such as the visual biofeedback tool for second language acquisition and silent speech interface. Unlike previous studies, here we explore the feasibility of age estimation using the ultrasound tongue image of the speakers. Motivated by the success of deep learning, this paper leverages deep learning on this task. We train a deep convolutional neural network model on the UltraSuite dataset. The deep model achieves mean absolute error (MAE) of 2.03 for the data from typically developing children, while MAE is 4.87 for the data from the children with speech sound disorders, which suggest that age estimation using ultrasound is more challenging for the children with speech sound disorder. The developed method can be used a tool to evaluate the performance of speech therapy sessions. It is also worthwhile to notice that, although we leverage the ultrasound tongue imaging for our study, the proposed methods may also be extended to other imaging modalities (e.g. MRI) to assist the studies on speech production.

\end{abstract}
\noindent\textbf{Index Terms}: age estimation, deep learning, ultrasound tongue imaging, convolutional neural network.

\section{Introduction}
During human natural speech, the analysis for the shapes and dynamics of the tongue is of great importance in the modeling of vocal tract \cite{maeda1990compensatory}. Ultrasound tongue imaging seems to be attractive, as the strength is that it images tongue motion at a fairly rapid frame rate (60 Hz or higher), which can capture subtle and swift movement during speech production \cite{Stone2005a}. Although some attempts have been made for 3D ultrasound tongue imaging \cite{naga2020automatic}, 2D B-mode ultrasound imaging is still widely-used to image the human tongue for over last three decades by phoneticians, predominately in a clinical context \cite{gick2012articulatory}. Compared with other imaging modalities (e.g. MRI), B-mode ultrasound imaging is non-invasive, convenient for experiment and less expensive than other imaging systems \cite{schwenzer2009non}. However, the Signal-to-Noise Ratio (SNR) of the ultrasound tongue image is quite low, and the speckle noise degrades the images by concealing fine structures and reduces the signal to noise level. Most of previous ultrasound tongue image analysis tasks focus on contour extraction \cite{Li2005b,Jaumard-Hakoun2015,xu2016robust,xu2016comparative}, articulatory-to-acoustic \cite{hueber2012continuous} or acoustic-to-articulatory Inversion \cite{porras2019dnn}.

It is widely known that speech signals contain both the dominant linguistic information and biophysical parameters, such as emotional state, health state, age, height \cite{kalluri2019deep}, ethnicity and identity of the speaker \cite{wen2019face}. Sustainable efforts \cite{ghahremani2018end} have been made to automatically extract such information from the speech signals, since the applications seem evident in practical scenarios, such as interactive voice response systems \cite{yacoub2003recognition}, age-dependent advertisements, caller-agent pairing and service customization.

In this paper, however, we are not aim to improve the age estimation performance using the audio signal, but using the ultrasound tongue imaging. Specifically, we endeavor to explore: whether the biophysical parameters of speakers (the age) can be inferred using the ultrasound tongue imaging. Generally speaking, even a trained clinical linguist cannot provide an accurate age estimation by just visual observing the ultrasound tongue image. It is incontrovertible that a person's voice is statistically related to the vocal tract structures. Directly relationship between the ultrasound tongue images and the ages of the speaker is that younger subjects generally image better than older subjects in most cases \cite{Stone2005a}. In the practical settings, the correlation between ultrasound tongue image and the age is complicated and hard to model, especially for the children with speech sound disorders \cite{eshky2019ultrasuite}. By providing an automatically accurate estimation can not only provide a tool to evaluate the speech therapy sessions, but also leads to the decreasing the workload for pathologists \cite{ribeiro2019speaker}.

Inspired by the success of deep learning, we explore whether deep learning approach can provide satisfied performance for the age estimation task using ultrasound tongue imaging. The rest of this paper is organized as follows: Sec. 2 briefly reviews the related literature; Sec. 3 presented our methodology. In Sec. 4, the experimental results are given. In Sec. 5, the conclusions are made and some potential future works are discussed.

\section{Relate work}
\subsection{Age estimation}
Due to the wide range of commercial applications, audio signal-based speaker's age estimation has recently received increased attention \cite{ghahremani2018end,bahari2012age}. However, speech signal-based age estimation is confronted with several challenges. (1) the perceived age of subject is usually different to the actual age (or chronological age); (2) a large annotated database with a wide, yet balanced, range of ages, is desirable to be obtained for developing a robust age estimation method. (3) the speech signals contain significant intra-speaker variability that is not closely correlated with the age \cite{bahari2012age}. And other factors, such as emotional condition, can also influence the speech signal and the age estimation performance. Recently the i-vector \cite{bahari2012age} and x-vector \cite{ghahremani2018end} based approach has been proposed, which obtained great performance for age estimation using the speech signals. However, to the best knowledge of the authors, there is no attempt to use ultrasound tongue image to predict the speaker's age yet.

\subsection{Ultrasound tongue image analysis using deep learning}
Ultrasound tongue image processing can be viewed as a sub-field of the medical ultrasound image processing. The traditional approach to quantify the motion of tongue is by extracting the tongue contours \cite{xu2016robust,li2005automatic} from the ultrasound image sequences. To date, deep learning has gained dramatically development for the ultrasound image analysis, while the potential applications of the deep learning in ultrasound tongue images are not well explored through some attempts have been made \cite{ribeiro2019speaker,zhu2018automatic,zhao2019predicting,xu2020predicting}. As a branch of machine learning, deep learning can be used for representation learning which can directly process and learning low-level, mid-level and high-level abstract features from the raw data. It holds the potential application of deep learning for ultrasound tongue image analysis tasks, such as tongue gestural classification \cite{ribeiro2019speaker}, contour extraction \cite{zhu2019cnn,hamed2019domain}, representation learning \cite{liu2016comparison,xu2017convolutional,xu2019ultrasound,li2019denoising}, cross-modality mapping \cite{porras2019dnn}. In this paper, we aim to explore a novel problem: whether it is possible to predict the age of the speaker using the ultrasound tongue image. However, the ultrasound tongue image-based age estimation is also quite challenge in the practical setting. For example, the ultrasound tongue image has quite low SNR, and the ultrasound imaging quality is not stable during the data recordings. Moreover, the size of the labeled dataset is much smaller with comparison to the labeled dataset of speech signals.

\section{Methodology}
\subsection{Data}
\begin{figure}[ht]
	\centering
	\includegraphics[scale=0.4]{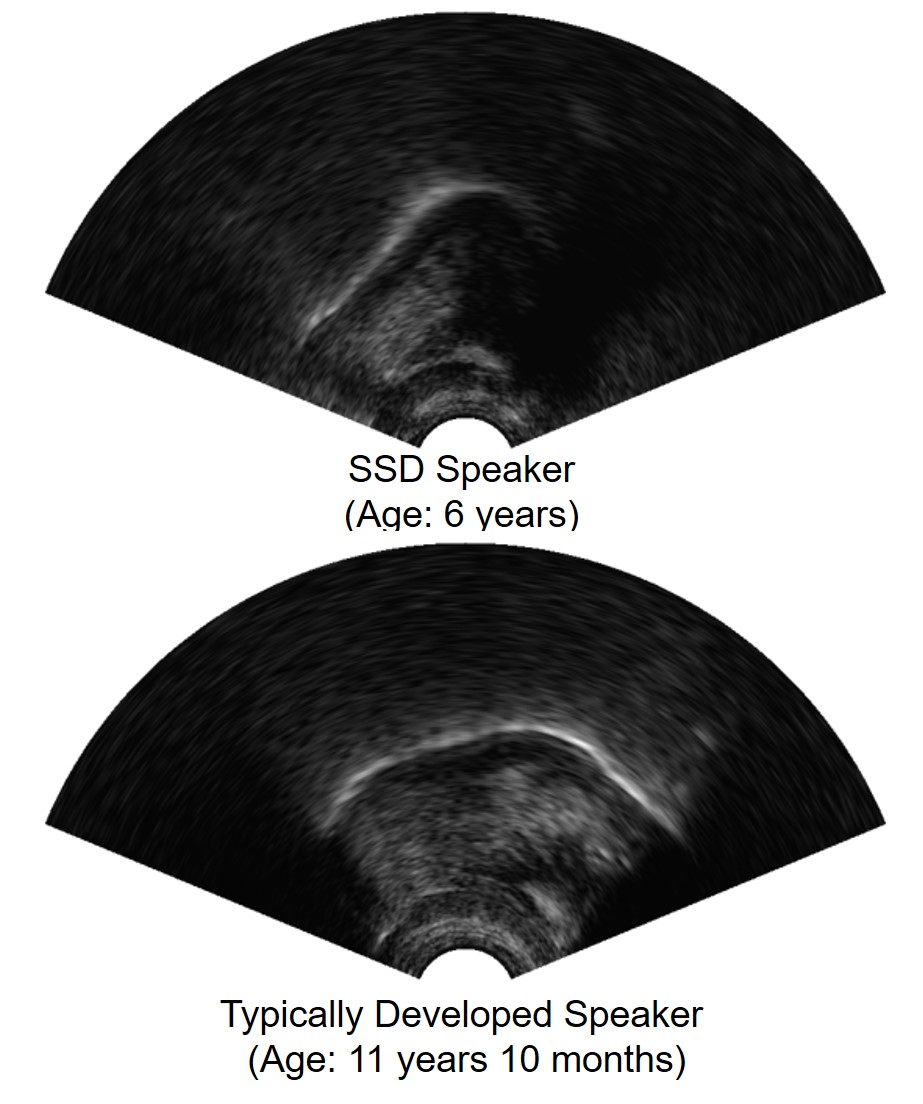}
	\caption{Sample frames from the dataset. The frame in the top row is extracted from UPX dataset, while the other frame in down row is from the UXTD dataset.}
	\label{sample_frame}
\end{figure}

In our experiments, we employ the dataset from the publicly available UltraSuite repository \cite{eshky2019ultrasuite}, which contains the ultrasound tongue imaging data recorded for typically developed children and the children with Speech Sound Disorders (SSD). Specifically, both UPX (with speech disorder) and UXTD (typical developed) datasets are used in our study and the detailed information of the dataset are given in Table 1, which includes the number of participants, the gender and ages information. As we aim to employ the ultrasound image for the age prediction task, we ignore the acoustic data and only leverage ultrasound images for the estimation task. The original size of single ultrasound frame 63$\times$412, which consists of 412 echo returns from each of the 63 scan lines. The sample frames are given in Figure ~\ref{sample_frame}. As depicted in the figure, it is quite difficult the estimate the speaker's age information by visual inspection.

\begin{table}[h]
	\centering 
	\caption{The detailed information of the dataset used in our study, which includes the number of participants, the gender and ages of the speakers.}
	\begin{tabular}{|c|c|c|}
		\hline
		& UPX & UXTD\\  
		\hline
		Number of participants & 20 & 58\\ 
		\hline
		Female & 4 & 31\\ 
		\hline
		Male  & 16 & 27\\ 
		\hline
		Mean age & 8y 4m & 9y 3m\\ 
		\hline
		Min age & 6y 1m & 5y 8m\\ 
		\hline
		Max age & 13y 4m & 12y 10m\\ 
		\hline
	\end{tabular}
	\label{dataset_information}
\end{table}

\subsection{Deep convolutional neural network}

Deep learning models achieve remarkable results in computer vision, natural language processing, and speech recognition. More attempts have been made to employ deep learning for ultrasound tongue imaging analysis \cite{zhu2018automatic,porras2019dnn,tatulli2017feature}. The well-known deep learning architectures include convolutional neural network, recurrent neural network, deep belief network and auto-encoders. Depending on the usage of label information, the deep learning models can be learned in either supervised or unsupervised manner. Here, we employ convolutional Neural Network (CNN) in our experiments, as CNN has a great potential to identify the various salient patterns of the images. The UltraSuite dataset gives the age information of the speaker, and we convert the age estimation task as supervised learning task using CNN.

There are numerous variants of CNN architectures in the literature \cite{gu2018recent}. However, for different CNN architectures been proposed, their basic components are very similar. Since the starting with LeNet \cite{lecun1998gradient}, CNNs have typically standard structure-stacked convolutional layers, optionally followed by batch normalization and max-pooling, and fully-connected layers. As deep learning is fast-grown field, we are not intent to explore the advanced neural network architectures. While, here we endeavor to provide a new perspective for the interpretation of the tongue using ultrasound. The details of the employed CNN architecture is given in Table \ref{cnn_information}. It is worthwhile to notice that the hyper-parameters are not optimized in our experiments, further optimization of the hyper-parameters may improve the performance on this task.

\begin{table}[h]
	\centering 
	\caption{The details of the CNN architectures used in this paper. The total number of trainable parameters is 289,8437.}
	\begin{tabular}{|c|c|c|}
		\hline
		Layer type & Output shape & Parameters\\ 
		\hline
		Input layer & (63, 412, 1) & \\  
		\hline
		2D Convolutional layer & (63, 412, 8) & 80\\ 
		\hline
		2D Convolutional layer & (61, 410, 8) & 584\\ 
		\hline
		2D Max pooling layer & (30, 205, 8) & 0\\ 
		\hline
		2D Convolutional layer &  (30, 205, 8) & 584\\ 
		\hline
		2D Convolutional layer &  (28, 203, 4) & 292\\ 
		\hline
		2D Max pooling layer & (14, 101, 4)  & 0\\ 
		\hline
		Dense connected layer & (512)  & 289,6384\\ 
		\hline
		Output layer & (1)  & 513\\ 
		\hline
	\end{tabular}
	\label{cnn_information}
\end{table}

\subsection{Real Time Data augmentation}
For different deep learning tasks, such as image classification, a lager number of annotated samples is critically for training, and more data may achieve higher performance by better generalization. The reason is that, when a model is trained with a small training set, the trained model tends to over-fit the training set and the model has poor generalization for the test data. Recently, data augmentation was widely used for the deep neural network, with the goal to virtually increase the amount of training data. Generally speaking, many data augmentation approaches apply a small mutation in the original training data and creating new samples artificially.
Here we use real time data augmentation to train the neural network, different augmentation have tested. We found the simple random rotation for the data augmentation achieves better performance with comparison to other different data augmentation approaches. The details can be found in the experimental results.

\section{Experimental results}

\subsection{Data Preparation}
For our experiments, we define our task as to learn a regression model for age estimation. For the UPX dataset, as each child may attend several different sessions \cite{eshky2019ultrasuite}, we employ the data from all sessions in our experiments. To represent each utterance, we select one frame from every 150 frames. Specifically, for the UXTD dataset, we have 24,449 frames in total. While for the UXTD dataset, we have 54,582 frames. We random select the training and validation datasets in our experiments. 80\% of the data is used for training, the remaining 20\% is used for validation.

\subsection{Training details}
To train the the CNN, we used Keras \footnote{https://keras.io/} python library with TensorFlow as the backend, which can fully utilize GPU resource. CUDA and cuDNN were also used to accelerate the learning system. A desktop computer with Windows 10 operation system with Intel 10-Core 3.7GHz CPU, 256 GB RAM, and 1 GeForce 2080Ti GPU is used in our experiment. RMSprop optimizer is used to compile the model with a customized learning rate of 0.001. To avoid the over-fitting during the training, a dropout strategy with a value of 0.5 is adopted for the fully connected layer. The ReLu activation function is used as the activation function for all of our experiments. The selected loss function and the evaluation metric is mean squared error (MSE), whose definition is given as follows:

\begin{equation}
MSE = \frac{1}{n}*\sum_{i=1}^{n} (Y_i - \hat{Y_i}) 
\end{equation}
where the $Y_i$ is the real age, while $\hat{Y_i}$ is predicted age using the trained deep neural network model. The maximum of the epoch is set as 100. The batch size is set as 128. The average time consumed is about 100 seconds for each epoch. It takes about 2-3 hours for the training process of each CNN.

\subsection{Ablation study for data augmentation}
As the size of the dataset is quite small, we found that data augmentation is vital to achieve better estimation performance. In this part, we conduct an experiment to test the influence of data augmentation for the age estimation task. We analyze the performance of CNN model across different data augmentation conditions, which include random rotation and adding random Gaussian noise with different parameters. Table~\ref{augmentation} shows the model's MSE with different data augmentation strategies. As can be seen from the table, random rotation can provide better performance in our experiments. The reason behind the is that: during the recording, small rotation may exist between the helmet and the ultrasound probe. In our following experiments, we use the random rotation augmentation strategies, with setting the maximum rotation angle as 5 degrees. 

\begin{table}[h]
	\centering 
	\caption{The performance of age estimation with different data augmentation strategies on the UXTD dataset. The parameters denotes the max value for the parameter during using the augmentation.}
	\begin{tabular}{|c|c|c|}
		\hline
		Augmentation strategies & Parameter & Validation MSE\\ 
		\hline
		No data augmentation  & / & 3.07 \\ 
		\hline
		Random rotation  & 5 &   \textbf{2.03}  \\  
		\hline
		Random rotation & 10 & 2.56\\ 
		\hline
		Random rotation & 15 & 2.63\\ 
		\hline
		Random rotation & 20 & 3.09\\ 
		\hline
		Random Gaussian noise & 0.01 & 2.96\\ 
		\hline
		Random Gaussian noise & 0.1  & 2.97\\ 
		\hline
		Random Gaussian noise & 0.2  & 3.34\\ 
		\hline
		Random Gaussian noise & 0.5  & 3.83\\ 
		\hline
	\end{tabular}
	\label{augmentation}
\end{table}

\subsection{Experiments Results for Typically Developing Children}
\begin{figure}[ht]
	\centering
	\includegraphics[scale=0.6]{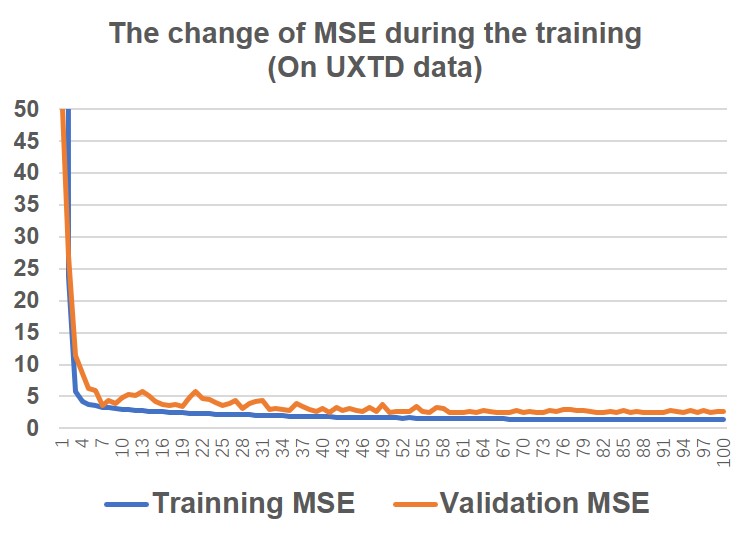}
	\caption{The change of loss function during the training phase on UXTD dataset. The vertical axis denotes the MSE, while the horizontal axis denotes the epoch number.}
	\label{cwssim}
\end{figure}

In this part, we report the performance of the trained CNN for the typically developing children. The change of loss function is given in Figure ~\ref{cwssim}. As can be seen from the figure, the MSE loss decreases rapidly in first 30 epochs. The min MSE on the validation data is about 2.03. Here, we use the mean age as the baseline, which provides a MSE of 3.64 on the validation data. In line with expectations, CNN provides a more accurate estimation than just using the average age. Based on the results, we suppose that the ultrasound tongue images contains the age information in the image, at least to some extent. On the other hand, the CNN does capture some patterns for age estimation by using the ultrasound tongue images as the input.

\subsection{Children with Speech Sound Disorder}

\begin{figure}[ht]
	\centering
	\includegraphics[scale=0.65]{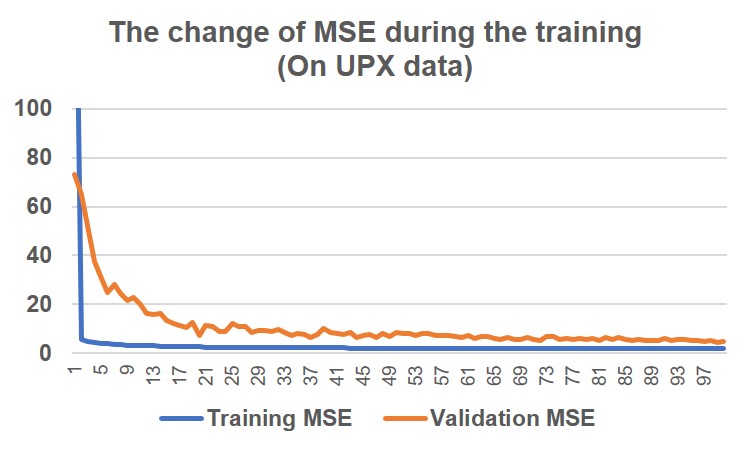}
	\caption{The change of loss function during the training phase on UPX dataset.}
	\label{SSD_Curve}
\end{figure}

We also conduct an experiment using the dataset from the children with speech sound disorders (SSD). The training curve is given in Figure \ref{SSD_Curve}. As can be seen from the figure, upon convergence, the model achieves a training MSE loss of 1.85 and 4.87 for validation MSE loss. The estimator using the average age can provide a MSE of 5.35.
The validation MSE is much larger with comparison to the one obtained using UXTD data. Also, the convergence speed is much slower. We suppose the reason behind these phenomenons include: due to the disorders, the children may repeating sounds and adding extra sounds during speech production. The children may also have difficulties to produce regular or complicated shapes during the data recordings. Thus, it is more challenging for the deep neural network to extract salient information for the age estimation task. 

\subsection{Model visualization}
In this part, we aim to visualize what does the CNN model learned for the age estimation task. Keract \footnote{https://github.com/philipperemy/keract} python library is used in our experiments.
We firstly visualize the convolutional layers in the learned convolutional neural networks. Similar to other image analysis tasks using CNN, the fist convolutions layers are used to extract low-level features from the image. As can be seen from the figures, the CNN model can extract the features from different gray scale (as shown in Figure 4). Secondly, we visualize the layers before the flatten layer. We observe that the CNN model not only focuses on the contour in the ultrasound tongue image, but also pay more attention to the regions correspondent to the tendon and tongue root regions, which may provide guidance for future ultrasound tongue imaging interpretation tasks.
\begin{figure}[ht]
	\centering
	\includegraphics[scale=0.45]{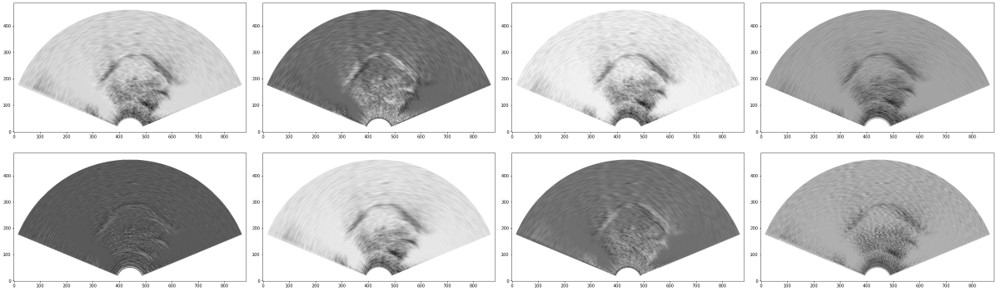}
	\caption{Visualization for the learned convolutional layers.}
	\label{TD_Curve}
\end{figure}

\begin{figure}[ht]
	\centering
	\includegraphics[scale=0.5]{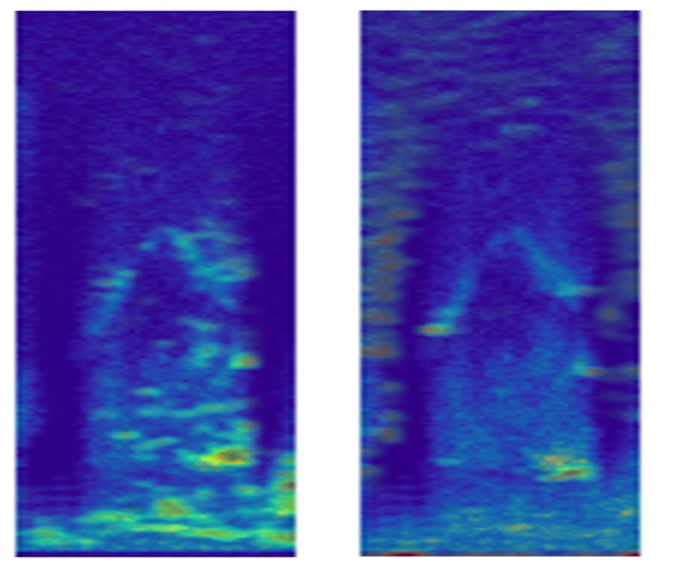}
	\caption{Heatmap for the layers learned in the CNN model (The last 2D max-pooling layer in the CNN network architecture). In the figure, the lowest activation are in dark, the highest are in dark red while the mid are in yellow.}
	\label{Heatmap}
\end{figure}

\section{Conclusions}
In this paper, we conducted a proof-of-concept study, which demonstrates that ultrasound tongue image can be used to predict the age of the speaker. To the best knowledge of the authors, this is the first attempt which aims to predict the age information using the ultrasound tongue image. This work demonstrates that deep learning techniques can be used improve the understanding the speech production process. On the other hand, our work also has several shortcomings:
(1). The ultrasound tongue imaging dataset is much smaller with comparison the speech signal-based dataset. Moreover, the range of the age is quite small. Thus, the estimation performance needs to be validated on larger dataset. 
(2). The performance is not optimal, as no optimization is conducted for the selection of the hyper-parameter and network architectures. 

For our further work, we would like to explore the performance of proposed method across different datasets and languages. It would also be interesting to extend our approach to distinguish the ultrasound tongue images recorded from the typical developed children or the children speech sound disorders. The current work may also be employed articulatory motion analysis.

\bibliographystyle{IEEEtran}
\bibliography{mybib}

\end{document}